\documentclass[runningheads]{llncs}
\usepackage{graphicx}
\usepackage{comment}
\usepackage{amsmath,amssymb} 
\usepackage{color}
\usepackage{graphicx}
\usepackage{subfigure}
\usepackage{multirow}

\usepackage[width=122mm,left=12mm,paperwidth=146mm,height=193mm,top=12mm,paperheight=217mm]{geometry}
\usepackage[colorlinks,linkcolor=red,anchorcolor=blue,citecolor=green]{hyperref}

\begin{document}
\pagestyle{headings}
\mainmatter
\def\ECCVSubNumber{2447}  

\title{Domain-Adaptive Few-Shot Learning} 

\titlerunning{Domain-Adaptive Few-Shot Learning}
%

\author{An Zhao$^*$\inst{1} \and
Mingyu Ding$^*$\inst{1,2} \and
Zhiwu Lu\thanks{Corresponding Author}\inst{1} \and
Tao Xiang\inst{3} \and
Yulei Niu\inst{1} \and
Jiechao Guan\inst{1} \and
Ji-Rong Wen\inst{1} \and
Ping Luo\inst{2}
\authorrunning{Zhao et al.}
%
\institute{
Beijing Key Laboratory of Big Data Management and Analysis Method \\
Gaoling School of Artificial Intelligence, Renmin University of China \and
The University of Hong Kong \and
Department of Electrical and Electronic Engineering, \\
University of Surrey, Guildford, Surrey GU2 7XH, United Kingdom \\
\email{zhaoan\_ruc@163.com, mingyuding@hku.hk, luzhiwu@ruc.edu.cn}}
}

\maketitle
\def\thefootnote{*}\footnotetext{Equal Contribution}

\begin{abstract}
  Existing few-shot learning (FSL) methods make the implicit assumption that the few target class samples are from the same domain as the source class samples. However, in practice this assumption is often invalid -- the target classes could come from a different domain. This poses an additional challenge of domain adaptation (DA) with few training samples. In this paper, the problem of domain-adaptive few-shot learning (DA-FSL) is tackled, which requires solving FSL and DA in a unified framework. To this end, we propose a novel domain-adversarial prototypical network (DAPN) model. It is designed to address a specific challenge in DA-FSL: the DA objective means that the source and target data distributions need to be aligned, typically through a shared domain-adaptive feature embedding space; but the FSL objective dictates that the target domain per class distribution must be different from that of any source domain class, meaning aligning the distributions across domains may harm the FSL performance.  How to achieve global domain distribution alignment whilst maintaining source/target per-class discriminativeness thus becomes the key. Our solution is to explicitly  enhance the source/target per-class separation before domain-adaptive feature embedding learning in the DAPN, in order to alleviate the negative effect of domain alignment on FSL. Extensive experiments show that our DAPN outperforms the state-of-the-art FSL and DA models, as well as their na\"ive combinations. The code is available at \href{https://github.com/dingmyu/DAPN}{https://github.com/dingmyu/DAPN}.
\end{abstract}

\section{Introduction}

Recently few-shot learning (FSL) \cite{fe2003bayesian,feifei2006pami,lake2013one} has received increasing interest. This is because, to scale a visual recognition model to thousands of (or even more) categories, one has to overcome the lack of labeled data problem. In particular, most visual recognition models are based on deep convolutional neural networks (CNNs). Training them typically requires hundreds of (or more) samples to be collected and annotated per class. This is often infeasible or even impossible for some rare categories. The goal of FSL is thus to recognize a set of target classes by learning with sufficient labelled samples from  source classes but only with a few labelled samples from the target classes.

\begin{figure}[t]
  \vspace{0.00in}
  \centering
  \includegraphics[width=0.70\columnwidth]{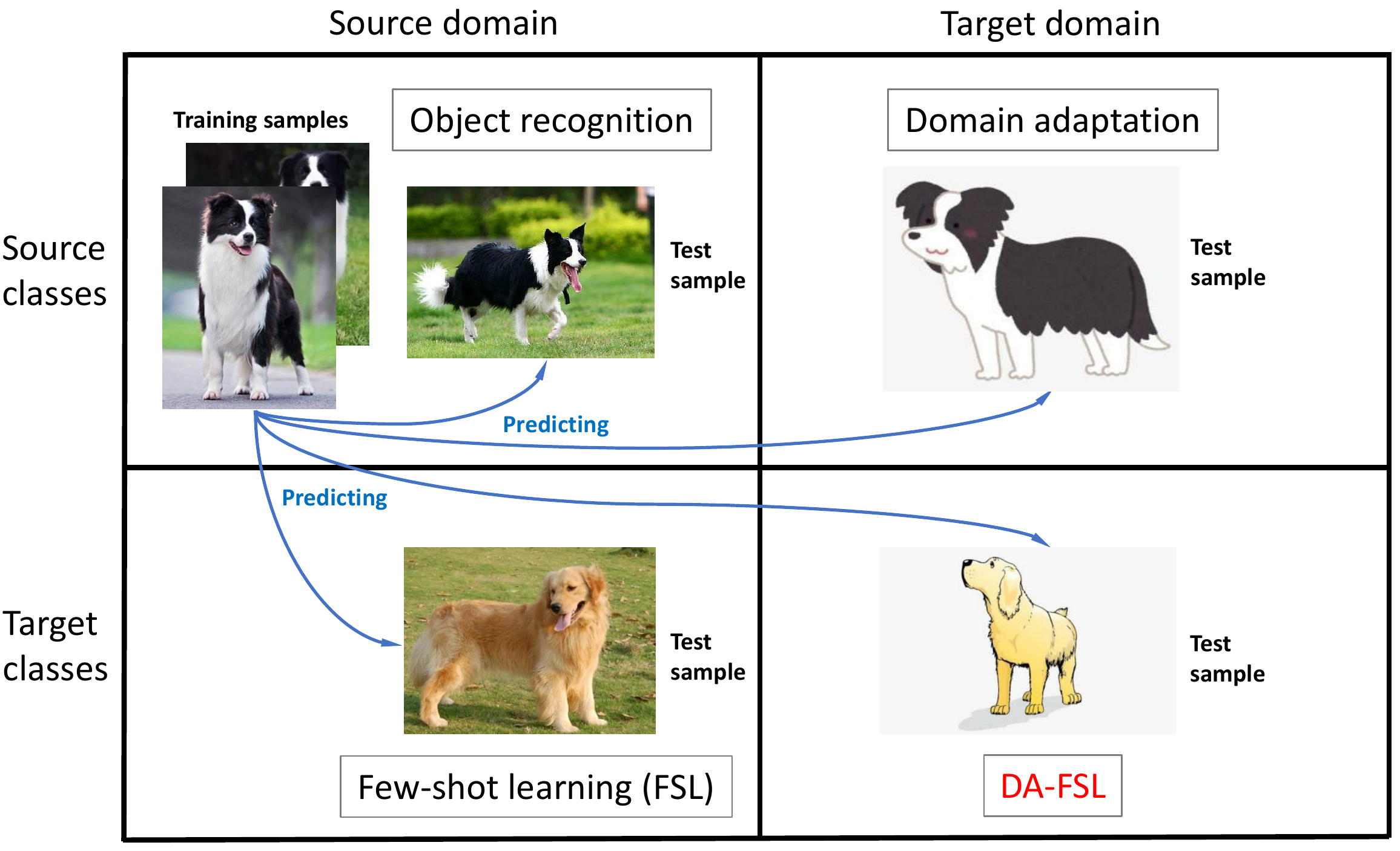}
  \vspace{-0.1cm}
  \caption{Illustration of the difference among four related visual recognition problems (i.e., many-shot objection recognition, FSL, domain adaptation, and DA-FSL).}
  \label{fig:problem}
  \vspace{-0.3cm}
\end{figure}

FSL \cite{NIPS2013_5209,Sun_2019_CVPR} is often formulated as a transfer learning problem \cite{pan2010tkde} from the source classes to the target ones. The efforts so far are mainly on how to build a classifier with few samples. However, there is an additional challenge that has largely been neglected so far, that is, the target classes  are not only poorly represented by the few training samples, but also can come from a different domain from that of the source classes. For example, the target class samples could be collected by a different imaging device (e.g. mobile phone camera vs.~single-lens reflex camera), resulting in different photo styles. In a more extreme case, the source classes could be captured in photos and the target ones in sketch or cartoon images. This means that the visual recognition model trained from the source classes needs to be adapted to both new classes and new domains,  with few samples from the target classes. This problem setting is termed as domain-adaptive few-shot learning (DA-FSL), which is illustrated in Fig.~\ref{fig:problem}.  

 DA-FSL is a more challenging problem due to the added objective of few-shot domain adaption. As far as we know, addressing both the few-shot DA and few-shot recognition problems jointly has never been attempted before. However, DA on its own, particularly unsupervised DA (UDA), has been studied intensively \cite{Tzeng2017cvpr,ganin2014unsupervised,bousmalis2016domain,long2018conditional,bousmalis2017unsupervised,Hoffman_cycada2017,xu2018unsupervised,rozantsev2019beyond,sun2016deep,kodirov2015unsupervised}. A straightforward solution seems to be combining a FSL with an existing DA method. In particular, most existing FSL methods \cite{snell2017prototypical,sung2018learning,ren2018meta,finn2017model} rely on feature reuse to the target classes in a feature embedding space learned from the source \cite{raghu2020rapid}. It is thus natural to introduce the DA learning objective by aligning the source and target data distributions in that embedding space.  Nevertheless, a na\"ive combination of existing DA and FSL methods fails to offer an effective solution (see Tables~\ref{tab_1shot}--\ref{tab_5shot}). This is because existing UDA methods assume that the target and source domains have identical label space. Given that they are mainly designed for distribution alignment across domains (recently focusing on per-class alignment \cite{saito2018maximum,luo2019taking,deng2019cluster,lee2019sliced,iccv19MME}), they are intrinsically unsuited for FSL whereby the target classes are completely different from the source ones: either global or per-class distribution alignment would have a detrimental effect on class separation and model discriminativeness. How to achieve domain distribution alignment for DA whilst maintaining source/target per-class discriminativeness thus becomes the key for DA-FSL.  
 
To this end, we propose a domain-adversarial prototypical network (DAPN) to solve the DA-FSL problem. Specifically, on top of the prototypical network \cite{snell2017prototypical} (designed for FSL), we introduce a novel adversarial learning method for few-shot domain adaptation. Note that domain adversarial learning has been popular among existing UDA methods \cite{ganin2014unsupervised,Hoffman_cycada2017,Tzeng2017cvpr,long2018conditional} for global (as opposed to per-class) distribution alignment. Since per-class alignment is the ultimate goal for UDA, its successful use in these UDA methods suggests that, global distribution alignment would indirectly lead to per-class alignment. That is an unwanted effect for our DA-FSL problem as the target classes are different from those of source. Therefore, in addition to the domain confusion objective commonly used by existing UDA methods for learning a domain-adaptive feature embedding space, new losses are introduced  before feature embedding (see Fig.~\ref{fig:framework}) to enforce source/target class discriminativeness. The end result is that we would have the better of both worlds: the global distributions of the source and target are aligned to reduce the domain gap for DA; in the meantime, the per-class distribution are not aligned and the source and target classes remain well-separable, benefiting the FSL task. With two sets of losses designed for DA and FSL respectively, to remove the need of weight selection for multiple losses, an adaptive re-weighting module is also introduced to further balance the two objectives.

Our contributions are: (1) The DA-FSL problem is formally defined and tackled. For the first time, we address both the few-shot DA and few-shot recognition problems jointly in a unified framework. (2) We propose a novel adversarial learning method to learn feature representation which is not only domain-confused for domain adaptation but also domain-specific for class separation. Extensive experiments show that our model outperforms the state-of-the-art FSL and domain adaptation models (as well as their na\"ive combinations).

\vspace{-0.2cm}
\section{Related Work}

\noindent\textbf{Few-Shot Learning}. In the past few years, FSL has been dominated by meta-learning based methods. They can be organized into three groups: (1) The first group adopts model-based learning strategies \cite{santoro2016one,munkhdalai2017meta} that fine-tune the model trained from the source classes and then quickly adapt it to the target classes. (2) The second group \cite{koch2015siamese,vinyals2016matching,snell2017prototypical,sung2018learning,ren2018meta} focuses on distance metric learning for the nearest neighbor (NN) search. Matching Network (MatchingNet) \cite{vinyals2016matching} builds different encoders for the support set and the query set. Prototypical Network (ProtoNet) \cite{snell2017prototypical} learns a metric space in which object classification can be performed by computing the distance of a test sample to the prototype representation of each target class. \cite{ren2018meta} makes improvements over ProtoNet towards a scenario where the unlabeled samples are also available within each episode. Relation Network (RelationNet) \cite{sung2018learning} recognizes the samples of new/target classes by computing relation scores between query images and the few samples of each new class. (3) The third group \cite{ravi2016optimization,finn2017model} chooses to utilize novel optimization algorithms instead of gradient descent to fit in the few-shot regime. \cite{ravi2016optimization} formulates an LSTM-based meta-learner model to learn an exact optimization algorithm used to train another neural network classifier in the few-shot regime. \cite{finn2017model} proposes a Model-Agnostic Meta-Learning (MAML) learner, whose weights are updated using the gradient, rather than a learned update rule. Although our DAPN model belongs to the second group with ProtoNet as a component, it is designed to address both few-shot DA and few-shot recognition problems (included in DA-FSL) jointly in a unified framework, which has not been studied before. 

\noindent\textbf{Domain Adaptation}. Note that the domain adaptation problem involved in our DA-FSL setting cannot be solved by supervised domain adaptation (SDA) \cite{motiian2017unified,abdelwahab2015supervised}. Although there exist a small set of labelled samples from the target domain used for DA under our DA-FSL setting, the classes from the target domain have no overlap with the classes from the source domain. Recently, unsupervised domain adaptation (UDA) has dominated the studies on DA. The conventional UDA models \cite{Fernando2013ICCV,gopalan2011domain,ni2013subspace,gong2012geodesic,tzeng2014deep,venkateswara2017deep,zhang2017joint,long2015learning,long2016unsupervised} typically leverage the subspace alignment technique. Many modern UDA methods \cite{Tzeng2017cvpr,ganin2014unsupervised,bousmalis2016domain,long2018conditional,bousmalis2017unsupervised,Hoffman_cycada2017,xu2018unsupervised,rozantsev2019beyond,sun2016deep,kodirov2015unsupervised} resort to adversarial learning \cite{goodfellow2014nips}, which minimizes the distance between the source and target features by a discriminator. However, as mentioned early, even if global domain distribution alignment is enforced, it often leads to per-class alignment which reduces the discriminativeness of the learned feature representation for the FSL task. Moreover, since existing UDA methods still assume that the target domain contains the same classes as the source domain, the more recent methods that focus on per-class cross-domain alignment \cite{saito2018maximum,luo2019taking,deng2019cluster,lee2019sliced,iccv19MME} are unsuitable for our DA-FSL problem. Global domain data distribution alignment \cite{Tzeng2017cvpr,laradji2018m,Hoffman_cycada2017} is thus adopted in our DAPN with special mechanism introduced to prevent per-class alignment. 

\noindent\textbf{Domain Adaptation + Few-Shot Learning}. Note that a cross-domain dataset (\emph{mini}ImageNet \cite{ravi2016optimization} $\rightarrow$ CUB \cite{CUB-200-2011}) is used for FSL in \cite{chen2019closerfewshot}. However, it is only for evaluating the cross-dataset generalization, rather than developing a new cross-domain FSL method. In contrast, this work focuses on much larger domain change (e.g. natural images vs. cartoon-like ones). Importantly, we develop a novel DA-FSL model to address the problem. Note that a new setting called few-shot domain adaptation (FSDA) is proposed \cite{motiian2017nips}. However, the FSDA setting in  \cite{motiian2017nips} is very different from ours in that: both source and target domains share the same set of classes under the FSDA setting, while the source and target classes have no overlap under our DA-FSL setting. \cite{sahoo2019metalearning} also proposes a DA-based FSL setting, but again it is very different from our work: in additional to a few labeled samples, \cite{sahoo2019metalearning} assumes the access to a large number of unlabeled samples from the target domain. In contrast, we do not make this assumption. Therefore, the problem setting in \cite{sahoo2019metalearning} is much easier than ours, and designed to exploit unlabeled target domain data, the method in \cite{sahoo2019metalearning}  cannot be used here.

\section{Methodology}
\label{method}

\subsection{Problem Definition}

Under our DA-FSL setting, we are given a large sample set $\mathcal{D}_s$ from a set of source classes $\mathcal{C}_s$ in a source domain, a few-shot sample set $\mathcal{D}_d$ from a set of target classes $\mathcal{C}_{d}$ in a target domain, and a test set $\mathcal{T}$ from another set of target classes $\mathcal{C}_t$ in the target domain, where $\mathcal{C}_s \cap \mathcal{C}_d = \emptyset$, $\mathcal{C}_t \cap \mathcal{C}_d = \emptyset$, and $\mathcal{C}_s \cap \mathcal{C}_t = \emptyset$. Our focus is then on training a model with $\mathcal{D}_s$ and $\mathcal{D}_d$ and then evaluating its generalization ability on $\mathcal{T}$. Note that there is also a few-shot sample set $\mathcal{D}_t$ (i.e. the support set) from the set of target classes $\mathcal{C}_{t}$, which could also be used for model training. However, we follow the FSL methods that do not require finetuning \cite{chen2019closerfewshot} and thus ignore $\mathcal{D}_t$ in the training phase. Due to the domain differences, the data distribution $P_s(x)$ for the set of source classes $\mathcal{C}_s$ is different from that (i.e. $P_t(x)$) for the set of target classes $\mathcal{C}_t\cup \mathcal{C}_{d}$, where $x$ denotes a sample. Formally, we have $\mathcal{D}_s =\{ (x_1, y_1), \dots, (x_N, y_N)~|~x_i \sim P_s(x), y_i \in \mathcal{C}_s \}$ and $\mathcal{D}_d = \{ (x_1, y_1), \dots, (x_K, y_K)~|~x_i \sim P_t(x), y_i \in \mathcal{C}_{d}\}$, where $y_i$ denotes the class label of sample $x_i$. The goal of our DA-FSL is to exploit $\mathcal{D}_s$ and $\mathcal{D}_d$ for training a classifier that can generalize well to $\mathcal{T}$. 

\begin{figure*}[t]
  \vspace{0.0in}
  \centering
  \includegraphics[width=0.9\textwidth]{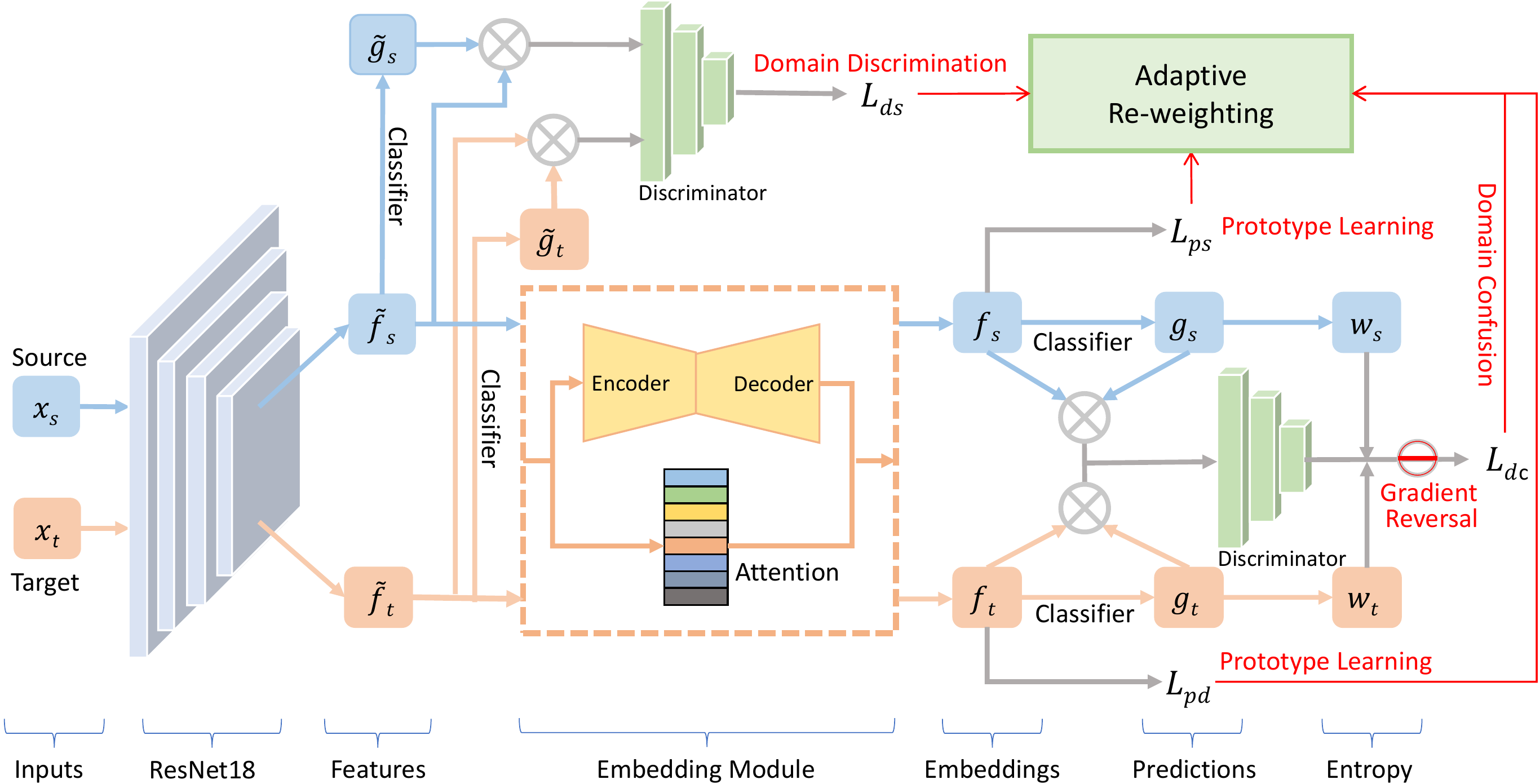}
  \vspace{0.0in}
  \caption{Overview of the proposed DAPN model for DA-FSL. Both source/target domain confusion and domain discrimination are explicitly included.  }
  \label{fig:framework}
  \vspace{-0.0in}
\end{figure*}

The proposed DAPN model is illustrated in Fig.~\ref{fig:framework}. Various modules in the network are designed for few-shot learning,  domain adaptation, as well as adaptive re-weighting to balance the two main objectives. They are introduced in details in the next three subsections respectively. 

\vspace{-0.3cm}
\subsection{Few-Shot Learning Module}
\label{method:fsl}

\vspace{-0.1cm}
\subsubsection{Episode Training}

To simulate the few-shot test process in the training phase, a small amount of data from both $\mathcal{D}_s$ and $\mathcal{D}_d$ are sampled to form episodic training sets. Specifically, we first build training episodes from the large sample set $\mathcal{D}_s$. To form a training episode $e_s$, we randomly choose $N_{sc}$ classes from $\mathcal{D}_s$ and then build two sets of samples from the $N_{sc}$ classes: the support set $S_s$ consists of $k \times N_{sc}$ samples ($k$ samples per class), and the query set $Q_s$ is composed of samples from the same $N_{sc}$ classes. For an $N_{meta}$-way $k$-shot problem, we train our model with an $N_{sc}$-way $k$-shot training episode, where $N_{sc} > N_{meta}$, as in \cite{vinyals2016matching,snell2017prototypical}. For example, if we perform $5$-way classification and $5$-shot learning in the test phase, each training episode could be generated with $N_{sc}=20$ and $k=5$. In addition to the training episodes from $\mathcal{D}_s$, we also build training episodes from the few-shot sample set $\mathcal{D}_d$. Since \textit{the samples in $\mathcal{D}_d$ are scarce} and even cannot form a single training episode, we perform the standard data augmentation method (i.e. horizontal flips and 5 random crops widely used for training existing CNN models) on $\mathcal{D}_d$, and obtain an augmented sample set $\hat{\mathcal{D}}_d$. To form a training episode $e_d$, we then randomly choose $N_{dc}$ classes from $\hat{\mathcal{D}}_d$ and build two sets of samples from the $N_{dc}$ classes: the support set $S_d$ contains $k \times N_{dc}$ samples with $k$ samples per class, and the query set $Q_d$ is sampled from remainder of the same $N_{dc}$ classes. In this work, we set $N_{dc} = N_{meta}$.

\vspace{-0.3cm}
\subsubsection{Prototypical Network}

Prototypical network \cite{snell2017prototypical} is selected as the main FSL component in our model because it is simple yet remains very competitive \cite{chen2019closerfewshot}. It learns a \emph{prototype} of each class in the support set $S_s$ and classifies each sample in the query set $Q_s$ based on the distances between each sample and different prototypes (i.e. the nearest neighbor classifier is used). Specifically, the $M$-dimensional prototypes are computed through an embedding function $f_\varphi: \mathcal{R}^d \to \mathcal{R}^M$ with learnable parameters $\varphi$. With the embedding function $f_\varphi$, the samples are projected from the $d$-dimensional visual  space into an $M$-dimensional feature space where the samples from the same class are close to each other and the samples from different classes are far away. 

Formally, the prototype $p^s_{c}$ of class $c$ in the support set $S_s$ is defined as the mean vector of the embedded support samples belonging to this class:
\begin{equation}
p^s_{c} = \frac{1}{|S_c|} \sum_{(x_i, y_i)\in S_c} f_\varphi(x_i),
\end{equation}
where $S_c = \{(x_i, y_i): (x_i, y_i)\in S_s, y_i=c\}$ denotes the set of support samples from class $c$.

Prototypical network then produces the class distribution of a query sample $x$ based on the softmax output w.r.t. the distance between the sample embedding $f_\varphi(x)$ and the class prototype $p^s_{c}$ as follows:
\begin{equation}
p_\varphi(y=c|x) = \frac{\exp(-\mathrm{dist}(f_\varphi(x), p^s_{c}))} {\sum_{c^\prime}{\exp(-\mathrm{dist}(f_\varphi(x), p^s_{c^\prime}))}},
\end{equation}
where $\mathrm{dist}(\cdot,\cdot)$ denotes the Euclidean distance in the $\mathcal{R}^M$ space. With the above class distribution, the loss function over each episode $e_s$ is defined based on the negative log-probability of query sample $x$ w.r.t. its true class label $c$:
\begin{equation}
L_{ps} = \mathbb{E}_{S_s,Q_s} [-\sum_{(x,y) \in Q_s} \log p_\varphi(y=c|x)].
\label{eq:source_fsl}
\end{equation}

Similarly, the loss function over each episode $e_d$ can be formulated based on the negative log-probability of query sample $x$ w.r.t. its true class label $c$:
\begin{equation}
L_{pd} = \mathbb{E}_{S_d,Q_d} [-\sum_{(x,y) \in Q_d} \log p_\varphi(y=c|x)]. \label{eq:loss_dam}
\end{equation}

The above two losses for prototype learning are employed in our proposed DAPN model on the feature output of a domain-adaptive embedding module (see Fig.~\ref{fig:framework}), which is described next. 

\vspace{-0.2cm}
\subsection{Domain Adversarial Adaptation Module}
\label{method:daa}

As mentioned before, the main objective of domain adaptive module is to learn a feature embedding space where the global distribution of the source and target domains are aligned, while the domain-specific discriminative information is still kept. To this end, we choose to enforce domain discriminativeness and domain alignment learning objectives before and after an embedding module. The task of balancing these two objectives are then handled by an adaptive loss re-weighting module to be described in Sec.~\ref{sec:reweighting}. 

\vspace{-0.3cm}
\subsubsection{Domain Adaptive Embedding} 

As shown in Fig.~\ref{fig:framework}, the input to the embedding module is the output of a feature extraction CNN (ResNet18 in this work), which represents each sample (image) $x$ as a 512-dimensional feature vector: $\mathrm{\tilde{f}} = \tilde{F}(x)$. The embedding module consists of an autoencoder and an attention sub-module. Concretely, the autoencoder takes $\mathrm{\tilde{f}}$ as input and output an embedding vector $\mathrm{\bar{f}}=\bar{F}(x)$. Moreover, to enforce $\mathrm{\bar{f}}$ to be as domain-confused as possible, we impose an attention sub-module composed of a fully-connected (FC) layer on it: the attention score  $\mathrm{sigmoid}(\mathrm{FC}(\mathrm{\tilde{f}}))$ is used to remove any domain-specific information (where $\mathrm{FC}(\cdot)$ denotes the output of the FC layer). Combining the autoencoder and attention sub-module together, we have the final output of the embedding module as $\mathrm{f}= F(x)$. 

\vspace{-0.3cm}
\subsubsection{Domain Adaptive Loss} 

Although both the autoencoder and attention sub-module can implicitly align the two domains, further alignment is needed by introducing domain adaptive losses.  Motivated by the superior performance of Conditional Domain Adversarial Network (CDAN) \cite{long2018conditional} on the domain adaptation task, we define a domain adversarial loss function $E$ on the domain discriminator $D$ across the source distribution $P_s(x)$ and target distribution $P_t(x)$, as well as on the feature representation $\mathrm{f}=F(x)$ after the feature embedding module and the classifier prediction $\mathrm{g}=G(x)$:
\begin{equation}
\begin{aligned}
 \min_{D}\max_{F,G}E =  -\mathbb{E}_{x_i^s \sim P_s(x)}\log[D(\mathrm{f}_i^s,\mathrm{g}_i^s)] -\mathbb{E}_{x_j^t \sim P_t(x)}\log[1-D(\mathrm{f_j}^t,\mathrm{g}_j^t)]. \label{eq:cdanloss0}
\end{aligned}
\end{equation}

Let $\mathrm{h}=(\mathrm{f},\mathrm{g})$ be the joint variable of feature representation $\mathrm{f}$ and classifier prediction $\mathrm{g}$. Concretely, the multilinear map $T_\otimes(\mathrm{h}) =\mathrm{f} \otimes \mathrm{g}$ is chosen to condition $D$ on $\mathrm{g}$, which is defined as the outer product of multiple random vectors. However, multilinear map faces dimension explosion. Let $d_\mathrm{f}$ and $d_\mathrm{g}$ denote the dimensions of vectors $\mathrm{f}$ and $\mathrm{g}$, respectively. The multilinear map has a dimension of $d_\mathrm{f} \times d_\mathrm{g}$, which is often too high dimensional to be embedded into deep learning models. To address this dimension explosion problem, the inner-product $T_\otimes(\mathrm{f},\mathrm{g})$ can be approximated by the dot-product $T_\odot(\mathrm{f},\mathrm{g})= \frac{1}{\sqrt{d}}(R_\mathrm{f} \mathrm{f})\odot(R_\mathrm{g} \mathrm{g})$, where $\odot$ is the element-wise product, $R_\mathrm{f} \in \mathcal{R}^{d\times d_f}$ and $R_\mathrm{g} \in \mathcal{R}^{d\times d_g}$ are two random matrices sampled only once and fixed in the training phase, and $d \ll d_\mathrm{f} \times d_\mathrm{g}$. Note that each element in $R_\mathrm{f}$ or $R_\mathrm{g}$ follows a symmetric distribution with invariance such as the uniform distribution and Gaussian distribution. Finally, we adopt the following conditioning strategy:
\begin{equation}
T(h) =
\begin{cases}
T_\otimes(\mathrm{f},\mathrm{g}) \quad  \mbox{if $d_\mathrm{f} \times d_\mathrm{g} \le d_{feat}$} \\
T_\odot(\mathrm{f},\mathrm{g}) \quad  \mbox{otherwise},
\end{cases}
\end{equation}
where $d_{feat}$ denotes the dimension of the output of the fully-connected layer. For domain adaptation, we solve an optimization problem derived from Eq.~(\ref{eq:cdanloss0}):
\begin{equation}
\begin{aligned}
\min_{D}\max_{T}E = -\mathbb{E}_{x^s_i \sim P_s(x)} \log [D(T(\mathrm{h}^s_i))]  - \mathbb{E}_{x^t_j \sim P_t(x)}\log [1-D(T(\mathrm{h}^t_j))],  \label{eq:cdanloss}
\end{aligned}
\end{equation}
where the subproblem of $\max_{T}E$ is solved by adding an gradient adversarial layer (see Fig.~\ref{fig:framework}) as in \cite{ganin2014unsupervised}, and the subproblem of $\min_{D}E$ is solved with the standard back propagation.

Note that some samples are easy-to-transfer, while others are hard-to-transfer. If the loss function imposes equal importance for different samples, it could weaken the effectiveness of the learned model. We thus modify the original CDAN \cite{long2018conditional} formulation by adopting the entropy criterion $H(\mathrm{g})=-\sum_{c=1}^{C}\mathrm{g}_c\log \mathrm{g}_c$, where $C$ is the number of classes and $g_c$ is the probability of the sample belong to class $c$. We re-weight training samples by an entropy-aware weight $w(H(\mathrm{g}))=1+e^{-H(\mathrm{g})}$ to make easy-to-transfer examples priority to hard ones. The loss for learning domain-confused feature representation is formulated as:
\begin{equation}
\begin{aligned}
L_{dc} = &  -\mathbb{E}_{x^s_i \sim P_s(x)} w(H(\mathrm{g}_i^s))\log [D(T(\mathrm{h}^s_i))] \\
& - \mathbb{E}_{x^t_j \sim P_t(x)} w(H(\mathrm{g}_j^t))\log [1-D(T(\mathrm{h}^t_j))].
\label{eq:cdaneloss}
\end{aligned}
\end{equation}

\vspace{-0.3cm}
\subsubsection{Domain Discriminative Loss} Note that the domain adaptive/confusion loss in Eq.~(\ref{eq:cdaneloss}) is useful for bridging the domain gap between source and target, but it also has the unwanted side-effect of over-alignment at per-class level which will harm the FSL performance. To alleviate this problem, we introduce a domain discrimination loss so that the per-class distributions within each domain is different from each other. Note that there is already a domain discriminator for domain alignment after embedding via gradient reversal (see Fig.~\ref{fig:framework}), so it makes little sense to add another on the same embedding space. Instead, our domain discriminative loss is added on the output of the feature extraction CNN.   

Concretely, we first define a conventional classification loss function $\tilde{E}$ on the domain discriminator $\tilde{D}$ across the source distribution $P_s(x)$ and target distribution $P_t(x)$, as well as on the feature representation $\mathrm{\tilde{f}}=\tilde{F}(x)$ before feature embedding and the classifier prediction $\mathrm{\tilde{g}}=\tilde{G}{(x)}$:
\begin{equation}
\begin{aligned}
\min_{\tilde{D},\tilde{F},\tilde{G}}\tilde{E}=  -\mathbb{E}_{x_i^s \sim P_s(x)}\log[\tilde{D}(\tilde{\mathrm{f}}_i^s,\mathrm{\tilde{g}}_i^s)]  -\mathbb{E}_{x_j^t \sim P_t(x)}\log[1-\tilde{D}(\tilde{\mathrm{f}}_j^t, \mathrm{\tilde{g}}_j^t)]. 
\label{eq:cdanloss1}
\end{aligned}
\end{equation}
Let $\mathrm{\tilde{h}}=(\mathrm{\tilde{f}},\mathrm{\tilde{g}})$. The loss for learning domain-specific feature representation is:
\begin{equation}
\begin{aligned}
L_{ds} =   -\mathbb{E}_{x^s_i \sim P_s(x)} \log [\tilde{D}(T(\tilde{\mathrm{h}}^s_i))] - \mathbb{E}_{x^t_j \sim P_t(x)} \log [1-\tilde{D}(T(\tilde{\mathrm{h}}^t_j))].
\label{eq:reversecdaneloss}
\end{aligned}
\end{equation}

\subsection{Adaptive Re-weighting Module}
\label{sec:reweighting}

Our DAPN model is trained with multiple objectives mentioned above (i.e. Eqs.~(\ref{eq:source_fsl}) (\ref{eq:loss_dam}) (\ref{eq:cdaneloss}) (\ref{eq:reversecdaneloss})), which can be viewed as multi-task learning. Among the losses, the FSL losses in Eqs.~(\ref{eq:source_fsl}) (\ref{eq:loss_dam}) and the domain discriminative loss in (\ref{eq:reversecdaneloss}) are pulling in different directions as the domain adaptive loss in (\ref{eq:cdaneloss}). This makes it more crucial to balance among them, especially since in different episodes, different recognition tasks are sampled which pose different level of demand for these competing learning objectives. A  na\"ive weighted sum of losses thus does not suffice. More sophisticated adaptive loss re-weighting mechanism is required. 

As reported in \cite{kendall2017multi}, there exists task-dependent uncertainty in multi-task learning, which stays constant for all input data and varies between different tasks. Therefore, we adopt an adaptive multi-task loss function based on maximizing the Gaussian likelihood with task-dependent uncertainty, in order to determine the weights of the objectives automatically.  Let the output of a neural network model with weights $\mathbf{W}$ on input $x$ be denoted as $\mathbf{f}^\mathbf{W}(x)$ (with $f^\mathbf{W}_c(x)$ be the $c$-th element of $\mathbf{f}^\mathbf{W}(x)$) and the discrete output of the model be denoted as $\mathrm{y}$. We utilize the classification likelihood to squash a scaled version of the model's output with a softmax function as follows:
\begin{equation}
p(\mathrm{y}|\mathbf{f}^\mathbf{W}(x)) = \text{softmax}(\mathbf{f}^\mathbf{W}(x)).
\end{equation}
Specifically, with a positive scalar $\sigma$, the log likelihood for this output is:
\begin{equation}
\begin{small}
\log p(\mathrm{y}=c|\mathrm{f}^\mathrm{W}(x), \sigma) = \frac{1}{\sigma ^ 2}f_c^\mathrm{W}(x) \hspace{-0.03in}-\hspace{-0.03in} \log \sum_{c'}
\exp (\frac{1}{\sigma ^ 2} f_{c'}^\mathrm{W}(x)). 
\end{small}
\end{equation}
In this work, our DAPN has four discrete outputs $\mathrm{y_1}, \mathrm{y_2}, \mathrm{y_3}, \mathrm{y_4}$, modeled with multiple softmax likelihoods, respectively. The joint loss $L(\mathbf{W}, \sigma_1, \sigma_2, \sigma_3, \sigma_4)$ is:
\begin{equation}
\begin{small}
\begin{aligned}
& L(\mathrm{W}, \sigma_1, \sigma_2, \sigma_3, \sigma_4) \\
& = \text{softmax}(\mathrm{y}_1\hspace{-0.03in} = \hspace{-0.03in} c;\mathrm{f}^\mathrm{W}(x), \sigma_1) \cdot \text{softmax}(\mathrm{y}_2 \hspace{-0.03in} = \hspace{-0.03in}c; \mathrm{f}^\mathrm{W}(x), \sigma_2) \\
& \hspace{0.14in}\cdot \text{softmax}(\mathrm{y}_3\hspace{-0.03in} = \hspace{-0.03in}c;\mathrm{f}^\mathrm{W}(x), \sigma_3) \cdot \text{softmax}(\mathrm{y}_4\hspace{-0.03in} = \hspace{-0.03in}c;\mathrm{f}^\mathrm{W}(x), \sigma_4)\\
& \approx \frac{1}{\sigma_1^2}L_1(\mathrm{W}) + \frac{1}{\sigma_2^2}L_2(\mathrm{W}) + \frac{1}{\sigma_3^2}L_3(\mathrm{W}) + \frac{1}{\sigma_4^2}L_4(\mathrm{W}) \\
& \hspace{0.14in} + \log\sigma_1 + \log\sigma_2 + \log\sigma_3 + \log\sigma_4. \nonumber
\end{aligned}
\end{small}
\end{equation}
In this paper, the adaptive weights among $L_1$, $L_2$, $L_3$ and $L_4$ are directly defined as: $w_j =\log\sigma_j^2~(j=1,2,3,4)$. Let $L_1=L_{ps}$~(see Eq.~(\ref{eq:source_fsl})), $L_2=L_{pd}$~(see Eq.~(\ref{eq:loss_dam})), $L_3=L_{dc}$~(see Eq.~(\ref{eq:cdaneloss})) and $L_4=L_{ds}$~(see Eq.~(\ref{eq:reversecdaneloss})). The overall loss of our model is thus formulated as follows:
\begin{equation}
\begin{small}
\begin{aligned}
& L = w_1/2 + \exp(-w_1)L_{s} + w_2/2 + \exp(-w_2)L_{d} \\
& + w_3/2 + \exp(-w_3)L_{dc} + w_4/2 + \exp(-w_4)L_{ds}.
\end{aligned}
\end{small}
\end{equation}

\section{Experiments}
\label{experiment}

\subsection{Datasets and Settings}

\begin{figure}[t]
\vspace{0.05in}
\centering
\includegraphics[width=0.9\columnwidth]{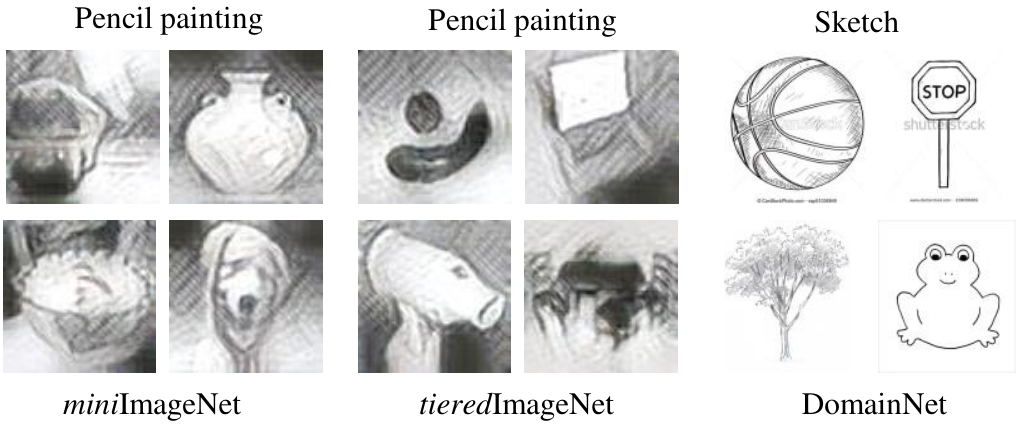}
\vspace{-0.0in}
\caption{Examples from the target domain for the three datasets. In each dataset, the source domain contains real/natural images. }
\label{fig:examples}
\vspace{-0.05in}
\end{figure}

\noindent\textbf{Datasets}. Three datasets are used for evaluation: (1) \textbf{\emph{mini}ImageNet} \cite{ravi2016optimization}: This dataset is a subset of ILSVRC-12 \cite{Russakovsky2015imagenet}. It consists of $100$ classes, and all images are of the size $84\times84$. We follow the widely-used class split as in \cite{ravi2016optimization} and adapt it to our DA-FSL setting: $64$ classes for $\mathcal{C}_s$ (with $600$ images per class), $16$ classes for $\mathcal{C}_d$ (with only $k$ images per class), and $20$ classes for $\mathcal{C}_t$ (with only $k$ labeled images per class to form the support set, and the other to form the test set). In this work, we set $k=1$ or 5. Further, we utilize the style transfer algorithm \cite{zhang2017multistyle} to transfer the samples from $\mathcal{C}_d$ and $\mathcal{C}_t$ into a new domain. Specifically, the samples of the source domain are natural pictures while the samples of the new/target domain are pencil paintings.
(2) \textbf{\emph{tiered}ImageNet} \cite{ren2018meta}: This dataset is also a subset of ILSVRC-12, but it is larger than \emph{mini}ImageNet. We use $351$ classes for $\mathcal{C}_s$ (with an average of $1,278$ images per class), $97$ classes for $\mathcal{C}_d$ (with only $k$ images per class), and $160$ classes for $\mathcal{C}_t$. All images are also of the size $84\times84$. The same style transfer is performed on the $\mathcal{C}_d$ and $\mathcal{C}_t$ splits of \emph{tiered}ImageNet to form a new domain.
(3) \textbf{DomainNet} \cite{peng2018moment}: To generate a new realistic dataset for DA-FSL, we exploit an existing multi-source domain adaptation dataset, which is the largest UDA dataset until now. There are $275$ classes for $\mathcal{C}_s$ (with an average of $516$ images per class), $55$ classes for $\mathcal{C}_d$ (with only $k$ images per class), and $70$ classes for $\mathcal{C}_t$. In this work, we take the real photo domain in DomainNet as the source domain and the sketch domain as the target domain. Each image is scaled to $84\times84$. For each of the above three datasets, examples from the target domain are shown in Fig.~\ref{fig:examples}.

\noindent\textbf{Evaluations}. We make evaluation on the test set under the 5-way 1-shot and 5-way 5-shot settings, as in previous works. The top-1 accuracy is computed for each test episode, and the average top-1 accuracy is reported over 2,000 test episodes (with $95\%$ confidence intervals).

\noindent\textbf{Baselines}. Three groups of baselines are selected: (1) \textbf{FSL Baselines}: Representative FSL baselines include relation network \cite{sung2018learning}, MatchingNet \cite{vinyals2016matching}, PPA \cite{Act2Param}, SGM \cite{yuxiongwang2017imaginary}, ProtoNet \cite{snell2017prototypical}, MetaOptNet \cite{lee2019meta} and Baseline++ \cite{chen2019closerfewshot}. We report the test results under the $5$-way $1$-shot and $5$-way $5$-shot settings. (2) \textbf{UDA Baselines}: Representative UDA baselines based on global domain-level alignment rather local class-level alignment are chosen. These include CDAN \cite{long2018conditional}, ADDA \cite{Tzeng2017cvpr}, AFN \cite{xu2018unsupervised}, M-ADDA \cite{laradji2018m}, and CyCADA \cite{Hoffman_cycada2017}. For testing under the 5-way 1-shot and 5-way 5-shot settings, we first train the CNN backbone with these UDA methods, and then extract the features of test/target samples so that a na\"ive nearest neighbor classifier can be used to recognize the test/target classes. (3) \textbf{UDA+FSL Baselines}: Representative baselines for directly combining UDA and FSL include CDAN+ProtoNet and CDAN+MetaOptNet (both trained end-to-end). We select the UDA+FSL baselines based on two criteria: 1) UDA baselines are latest/state-of-the-art (e.g. CDAN \cite{long2018conditional} is state-of-the-art); 2) FSL baselines are representative/state-of-the-art (e.g. ProtoNet \cite{snell2017prototypical} is representative and MetaOptNet \cite{lee2019meta} is state-of-the-art).

\noindent\textbf{Implementation Details}. Our model is implemented in PyTorch. The ResNet18 model \cite{DBLP:conf/cvpr/HeZRS16} is used as the backbone for all compared methods. We pretrain the backbone from scratch using the training set and then finetune it to solve the DA-FSL problem. In this work, the end-to-end training process is implemented by using back-propagation and stochastic gradient descent. The learning rate is initially set to $\eta_0=0.001$, and then is adjusted (as in \cite{long2018conditional}) by $\eta_p=\eta_0(1+\alpha p)^{-\beta}$, where $\alpha=10$, $\beta=0.75$, and $p$ is the training progress ranging from $0$ to $1$. A momentum of 0.9 and a weight decay of 0.01 are also selected for training. The code and datasets will be released soon. 
 
\begin{table}[t]
  \vspace{0.05in}
  \centering
  \caption{Comparative accuracies (\%, top-1) with 95\% confidence intervals under the DA-FSL setting (5-way 1-shot) on the three datasets.  }
  \tabcolsep6.5pt
  \scalebox{0.85}{
  \begin{tabular}{l|c|c|c}
    \hline
    Model  &  \emph{mini}ImageNet & \emph{tiered}ImageNet &  DomainNet\\
    \hline
    ADDA \cite{Tzeng2017cvpr} & $22.83\pm0.26$ & $25.31\pm0.31$ & $31.14\pm0.36$ \\
    CyCADA \cite{Hoffman_cycada2017} &$22.65\pm0.28$ & $25.28\pm0.33$ &$32.27\pm0.34$ \\
    AFN \cite{xu2018unsupervised} &$23.83\pm0.22$ & $25.74\pm0.24$ & $32.78\pm0.31$ \\
    CDAN \cite{long2018conditional} &$23.82\pm0.24$ & $25.82\pm0.30$ &$33.55\pm0.35$ \\
    M-ADDA \cite{laradji2018m} &$23.54\pm0.29$ & $25.92\pm0.32$ &$31.71\pm0.35$ \\
    \hline
    RelationNet \cite{sung2018learning} &$23.87\pm0.82$  & $24.12\pm0.84$ &$31.98\pm0.72$    \\
    MatchingNet \cite{vinyals2016matching} &$23.35\pm0.64$ & $25.53\pm0.46$ &$32.10\pm0.73$ \\
    PPA \cite{Act2Param}  &$23.86\pm0.42$ & $ 24.62\pm0.41 $ &$33.71\pm0.41$  \\
    SGM \cite{yuxiongwang2017imaginary}  &$23.49\pm0.29$ &$24.03\pm0.26$ &$ 33.29\pm0.27 $  \\
    ProtoNet \cite{snell2017prototypical} &$ 23.23\pm0.32 $ & $23.54\pm0.33$ &$33.66\pm0.36$  \\
    MetaOptNet \cite{lee2019meta}  & $ 24.53\pm0.20 $ & $ 25.06\pm0.33 $ &$ 34.50\pm0.36 $     \\
    Baseline++ \cite{chen2019closerfewshot} &$24.06\pm0.46$  &$24.65\pm0.74$ &$34.34\pm0.77$ \\
    \hline
    CDAN+ProtoNet &$25.36 \pm 0.21$ &$ 26.52\pm 0.23$ &$35.10 \pm 0.42$ \\ 
    CDAN+MetaOptNet &$ 25.78\pm0.23 $ &$ 26.87\pm0.41 $ &$ 35.46\pm0.36 $ \\ 
    \hline
    DAPN (ours) &$\mathbf{27.25}\pm0.25$ & $\mathbf{28.47}\pm0.25$ &$\mathbf{36.96}\pm0.35$ \\
    \hline
  \end{tabular}}
  \label{tab_1shot}
  \vspace{-0.0cm}
\end{table}

\begin{table}[t]
  \vspace{-0.0cm}
  \centering
  \caption{Comparative accuracies (\%, top-1) with 95\% confidence intervals under the DA-FSL setting (5-way 5-shot) on the three datasets. }
  \tabcolsep6.5pt
  \scalebox{0.85}{
  \begin{tabular}{l|c|c|c}
    \hline
    Model  &  \emph{mini}ImageNet & \emph{tiered}ImageNet &  DomainNet\\
    \hline
    ADDA \cite{Tzeng2017cvpr}  &$29.13\pm0.43$ & $30.22\pm0.44$  & $45.86\pm0.48$ \\
    CyCADA \cite{Hoffman_cycada2017}  &$29.36\pm0.33$ &$32.14\pm0.33$  &$48.11\pm0.52$\\
    AFN \cite{xu2018unsupervised}  &$32.56\pm0.30$ & $33.06\pm0.39$  & $50.22\pm0.49$ \\
    CDAN \cite{long2018conditional}  &$31.77\pm0.28$ & $34.11\pm0.31$ & $51.56\pm0.34$ \\
    M-ADDA \cite{laradji2018m} &$30.30\pm0.23$ &$33.56\pm0.33$  &$47.23\pm0.39$\\
    \hline
    RelationNet \cite{sung2018learning}  &$33.29\pm0.96$  &$33.15\pm0.94$ &$51.12\pm0.58$    \\
    MatchingNet \cite{vinyals2016matching}  &$32.42\pm0.55$ &$32.59\pm0.46$  &$51.07\pm0.74$    \\
    PPA \cite{Act2Param} &$33.74\pm0.41$  &$ 33.65\pm0.52 $  &$ 51.66\pm0.42 $    \\
    SGM \cite{yuxiongwang2017imaginary}   &$32.67\pm0.32$ & $33.42\pm0.31$  &$ 51.42\pm0.24 $  \\
    ProtoNet \cite{snell2017prototypical} &$ 32.92\pm0.41 $   &$33.38\pm0.29$  &$51.72\pm0.34$  \\
    MetaOptNet \cite{lee2019meta}  &$ 33.23\pm 0.63$  &$ 34.36\pm 0.25$  &$ 51.76\pm0.52 $    \\
    Baseline++ \cite{chen2019closerfewshot} &$32.74\pm0.81$   &$34.29\pm1.09$ &$51.73\pm0.70$  \\
    \hline
    CDAN+ProtoNet  &$35.51 \pm 0.25$ &$37.43 \pm 0.29$  &$52.10 \pm 0.42$  \\ 
    CDAN+MetaOptNet &$ 35.87\pm0.25 $ &$ 37.79\pm0.32 $ &$ 52.72\pm0.41 $  \\ 
    \hline
    DAPN (ours)  &$\mathbf{37.45}\pm0.25$ &$\mathbf{39.90}\pm0.29$ &$\mathbf{54.32}\pm0.36$ \\
    \hline
  \end{tabular}}
  \label{tab_5shot}
\vspace{-0.0cm}
\end{table}

\vspace{-0.1cm}
\subsection{Main Results}

The comparative results under our DA-FSL setting on the three datasets are shown in Tables~\ref{tab_1shot} and \ref{tab_5shot}. We have the following observations: (1) On all datasets, our DAPN  significantly outperforms the state-of-the-art FSL and UDA methods, because of its ability to tackle both problems. (2) Our DAPN model also clearly performs better than the two UDA+FSL baselines, showing that the na\"ive combination of UDA and FSL is not as effective as our specifically designed DAPN model for DA-FSL. (3) Interestingly, when combined with a na\"ive nearest neighbor classifier (for FSL), the performance of existing UDA methods is as good as that of any existing FSL methods. This suggests that solving the domain adaptation problem is the key for our DA-FSL setting. (4) Given the same 5-way 5-shot (or 5-way 1-shot) evaluation setting, the test results on the first two datasets are clearly worse than those on DomainNet. This indicates that the domain gap (induced by style transfer) and the category gap (induced by FSL) of the first two datasets are even bigger than those of the widely-used realistic dataset -- DomainNet. This justifies the inclusion of these two synthesized datasets for performance evaluation under the DA-FSL setting. 

\vspace{-0.1cm}
\subsection{Further Evaluations}

\begin{figure}[t]
\vspace{0.05in}
\centering
\subfigure[Ablation study for our full model]{
\includegraphics[width=0.35\linewidth]{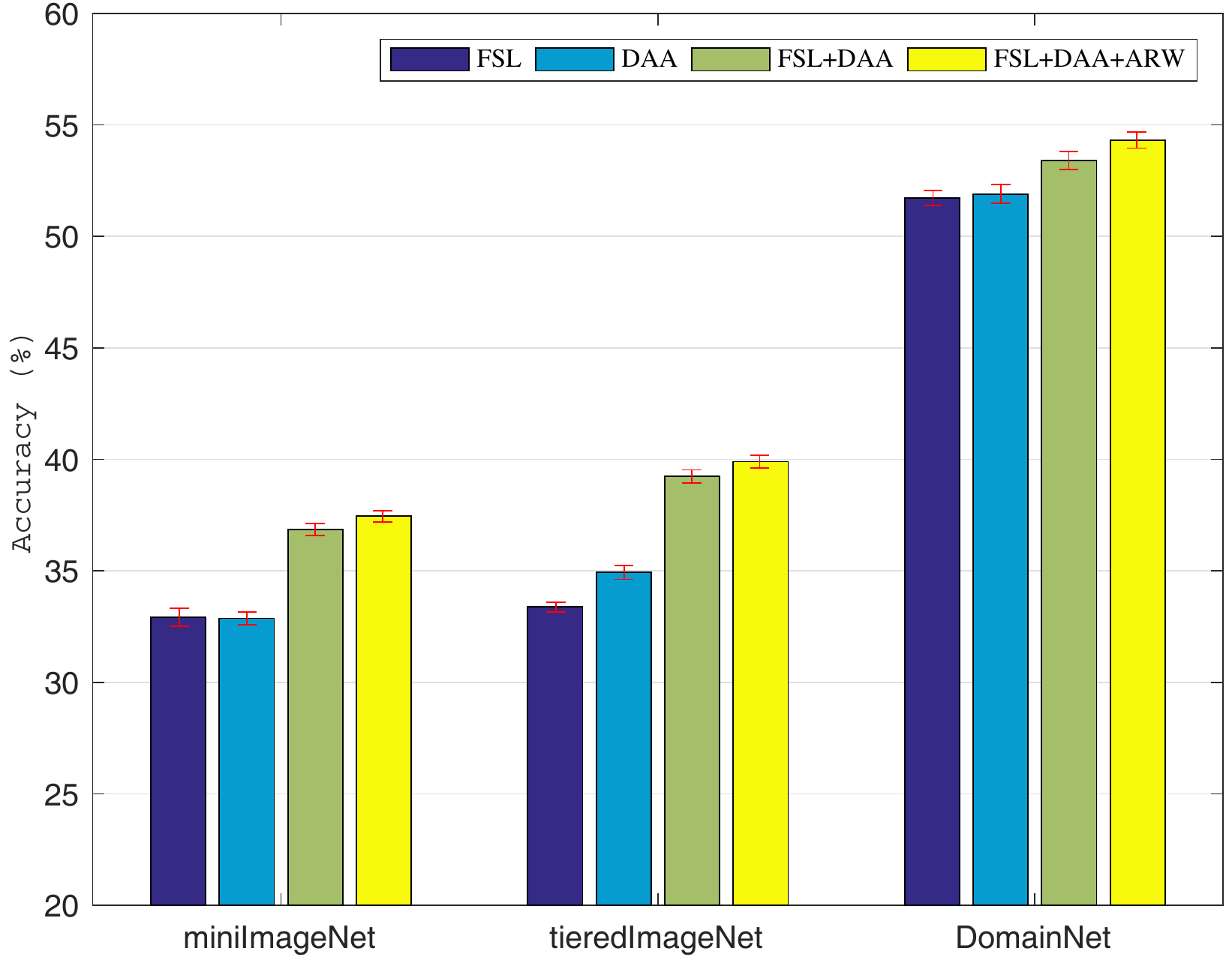}
\label{fig:ablation1}
}~~~~~~~~
\subfigure[Ablation study for our DAA module]{
\includegraphics[width=0.35\linewidth]{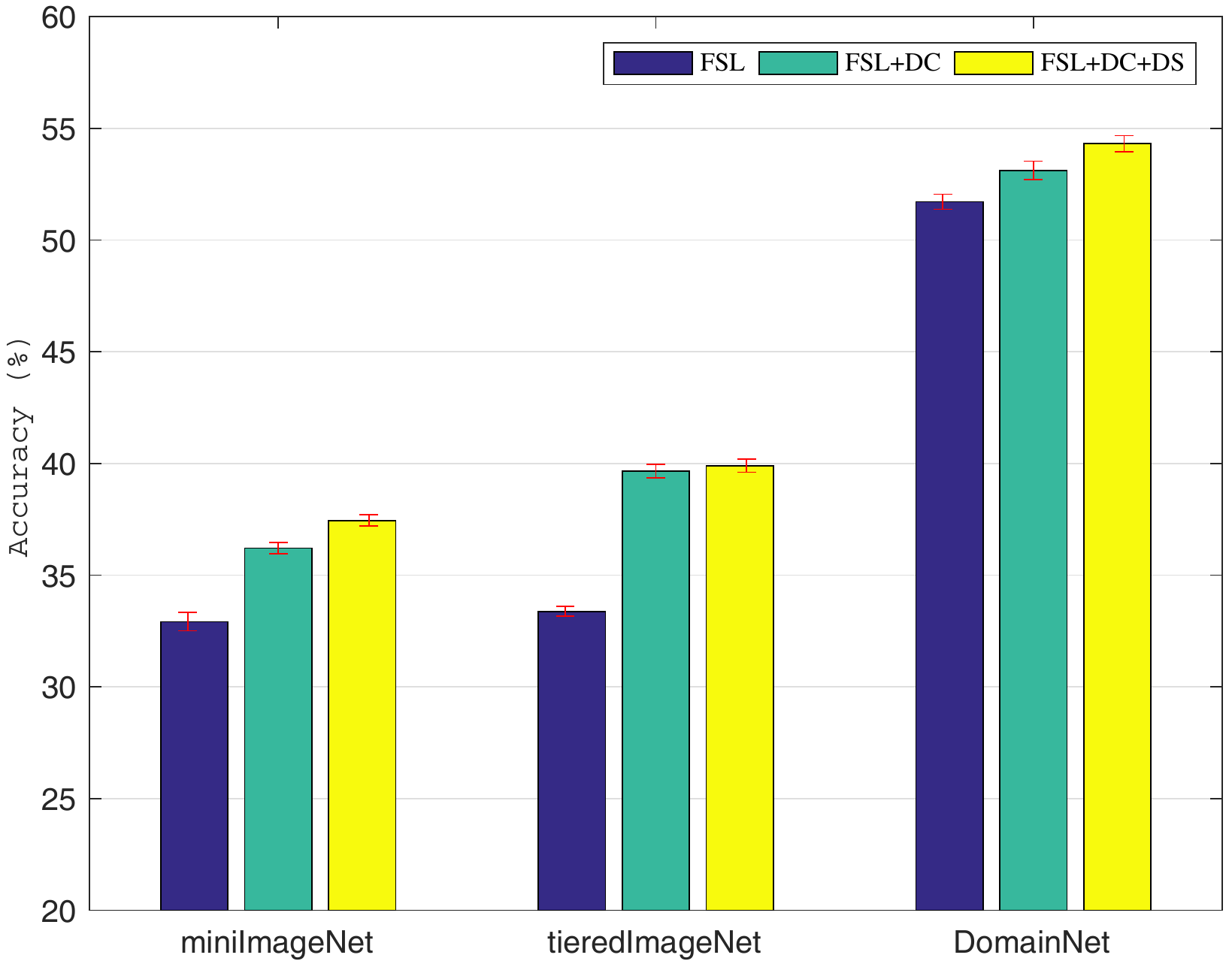}
\label{fig:ablation2}
}
\vspace{-0.1cm}
\caption{(a) Ablation study results for our full model under the DA-FSL setting (5-way 5-shot) on the three datasets; (b) Ablation study results for our DAA module under the DA-FSL setting (5-way 5-shot) on the three datasets. The error-bars show the $95\%$ confidence intervals. }
\label{fig:ablation}
\vspace{-0.1cm}
\end{figure}

\noindent\textbf{Ablation Study on Our Full Model}. To demonstrate the contribution of each module of our full DAPN model, we make comparison to its three simplified versions: (1) FSL -- only the few-shot learning (FSL) module (described in Section~\ref{method:fsl}) is used; (2) DAA -- the domain adversarial adaptation (DAA) module (described in Section~\ref{method:daa}) is combined with a na\"ive nearest neighbor classifier; (3) FSL+DAA -- the FSL and DAA modules are combined for DA-FSL without using adaptive re-weighting. Since our full model combines the two main modules using adaptive re-weighting (ARW), it can be denoted as Full or FSL+DAA+ARW. The ablation study is performed under the 5-way 5-shot DA-FSL setting. The obtained ablative results are presented in Fig.~\ref{fig:ablation1}. It can be seen that: (1) The performance continuously increases when more modules are used to solve the DA-FSL problem, demonstrating the contribution of each module. (2) The improvements achieved by DAA over the classical FSL suggest that the domain adaptation module is important for the DA-FSL setting and it can perform well even with the na\"ive nearest neighbor classifier. (3) The ARW module clearly yields performance improvements, validating its effectiveness in determining the weights of multiple losses.

\noindent\textbf{Ablation Study on Our DAA Module}. We further conduct ablation study to show the contribution of each component of our DAA module. Three methods are compared: (1) FSL -- FSL using the two losses $L_{ps}$ defined in Eq.~(\ref{eq:source_fsl}) and $L_{pd}$ defined in Eq.~(\ref{eq:loss_dam}); (2) FSL+DC -- DA-FSL using the three losses $L_{ps}$, $L_{pd}$, and $L_{dc}$ defined in Eq.~(\ref{eq:cdaneloss}); (3) FSL+DC+DS -- DA-FSL using the four losses $L_{ps}$, $L_{pd}$, $L_{dc}$, and $L_{ds}$ defined in Eq.~(\ref{eq:reversecdaneloss}). For fair comparison, adaptive re-weighting is used for all three methods. The ablative results on the three datasets are shown in Fig.~\ref{fig:ablation2}. We have two observations: (1) The significant improvements achieved by FSL+DC over FSL show that domain confusion after the embedding module is extremely important for our DA-FSL setting. (2) FSL+DC+DS consistently outperforms FSL+DC, validating the effectiveness of domain discrimination before the embedding module. 

\begin{figure}[t]
\vspace{0.05in}
\centering
\subfigure[DC: before embedding]{
\includegraphics[width=0.4\linewidth]{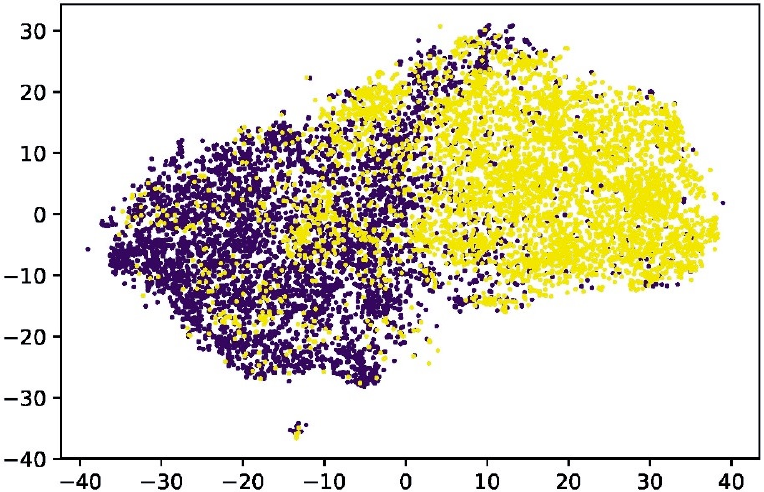}
\label{fig:tsne_daa1}
}%
\subfigure[DC: after embedding]{
\includegraphics[width=0.4\linewidth]{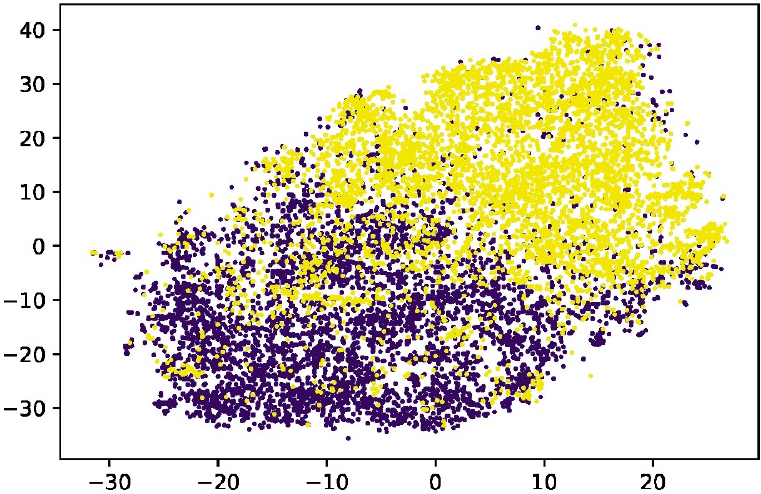}
\label{fig:tsne_daa2}
}
\subfigure[DC+DS: before embedding]{
\includegraphics[width=0.4\linewidth]{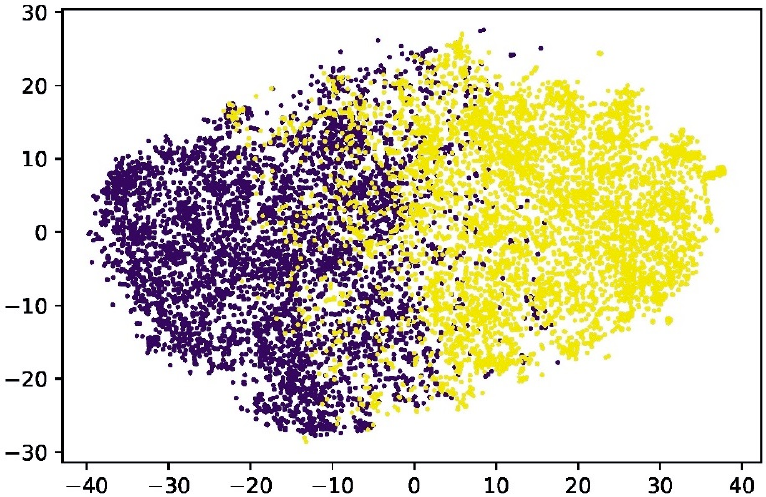}
\label{fig:tsne_daa3}
}%
\subfigure[DC+DS: after embedding]{
\includegraphics[width=0.4\linewidth]{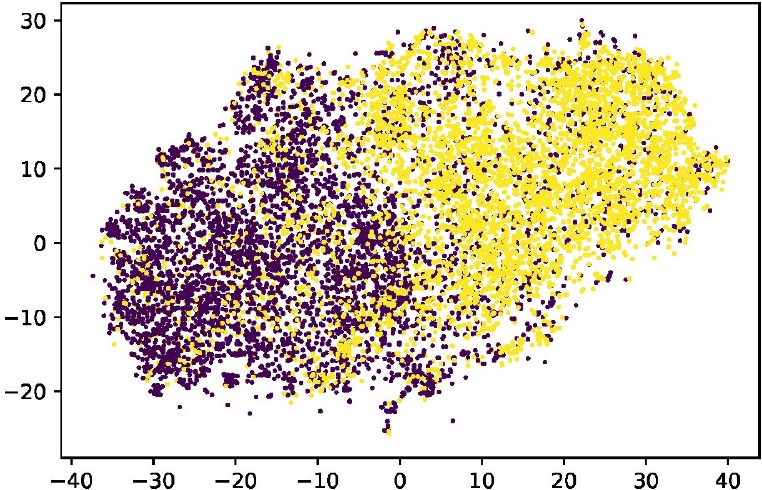}
\label{fig:tsne_daa4}
}
\vspace{-0.1cm}
\caption{The t-SNE visualization of the feature vectors of 5,000 randomly-selected images from the source domain (purple dots) and 5,000 images from the target domain (yellow dots) on the DomainNet dataset. \textit{Left}: feature vectors extracted before the embedding module; \textit{Right}: feature vectors extracted after the embedding module. Notations: DC -- domain adaptation using the loss $L_{dc}$ defined in Eq.~(\ref{eq:cdaneloss}); DC+DS -- domain adaptation using the loss $L_{dc}$ as well as the loss $L_{ds}$ defined in Eq.~(\ref{eq:reversecdaneloss}). }
\label{fig:feature_tsne}
\vspace{-0.0cm}
\end{figure}

\noindent\textbf{Feature Visualization for Our DAA Module}. The ablation study results shown in Fig.~\ref{fig:ablation2} are also supported by the t-SNE visualization of the feature vectors extracted before/after the embedding module. Some qualitative results can be seen in Fig.~\ref{fig:feature_tsne}. It shows that the addition of $L_{ds}$ (defined in Eq.~(\ref{eq:reversecdaneloss})) leads to two improvements: (1) The source/target samples are discriminated significantly better before embedding (see Fig.~\ref{fig:tsne_daa3} vs. Fig.~\ref{fig:tsne_daa1}); (2) The source/target samples are enforced to be more confused after embedding (see Fig.~\ref{fig:tsne_daa4} vs. Fig.~\ref{fig:tsne_daa2}). This explains the better performance of our DAA module (w.r.t. the conventional domain confusion) shown in Fig.~\ref{fig:ablation2}.

\section{Conclusion}

In this work, we have investigated a new FSL setting called DA-FSL. To simultaneously  learn a classifier for new classes with a few shots and bridge the domain gap, we proposed a novel DAPN model by integrating prototypical metric learning and domain adaptation within a unified framework. The domain discriminative and domain confusion learning objectives are introduced before and after a domain-adaptive embedding module, which are further balanced with an adaptive re-weighting module.  Extensive experiments showed that our DAPN model outperforms the state-of-the-art FSL and domain adaptation models. 

\clearpage

%
%

\bibliographystyle{splncs04}
\bibliography{dapn_eccv2020}
\end{document}